%% file: root.tex
\documentclass[letterpaper, 10 pt, conference]{ieeeconf}  % Comment this line out if you need a4paper
\usepackage{graphicx}
\usepackage{hyperref}

\usepackage{pdfpages}

\IEEEoverridecommandlockouts                              % This command is only needed if 
\usepackage{algorithm}
\usepackage{algorithmic}
\usepackage{amsmath}
\usepackage{multirow}
\usepackage{graphicx}

\title{\LARGE \bf
Robotic Automation in Apparel Manufacturing: A Novel Approach to Fabric Handling and Sewing}

\author{Abhiroop Ajith\textsuperscript{*}$^{1}$, Gokul Narayanan\textsuperscript{*}$^{1}$, Jonathan Zornow$^{2}$, Carlos Calle$^{3}$, Auralis Herrero Lugo$^{4}$, \\ Jose Luis Susa Rincon$^{1}$,   Chengtao Wen$^{1}$ and Eugen Solowjow$^{1}$% <-this % stops a space
\thanks{$^{*}$ These authors contributed equally to this work.}
\thanks{$^{1}$Abhiroop Ajith, Gokul Narayanan, Jose Luis Susa Rincon, Eugen Solowjow, Chengtao Wen are with Siemens Corporation
        {\tt\small \{abhiroop.ajith, gokul.sathya\_narayanan,
        eugen.solowjow,chengtao.wen\}@siemens.com}}%
\thanks{$^{2}$ Jonathan Zornow is with Sewbo.
    {\tt\small \{jon@sewbo.com\}}}%
\thanks{$^{3}$Carlos Calle is with Levi's. 
    {\tt\small \{ccalle@levi.com\}}}%
\thanks{$^{4}$Auralis Herrero Lugo is with Bluewater Defense.
        {\tt\small \{aherrero@bwdefense.com\}}}%
}    

\input{macros.tex}

\begin{document}

\maketitle
\thispagestyle{empty}
\pagestyle{empty}

%%%%%%%%%%%%%%%%%%%%%%%%%%%%%%%%%%%%%%%%%%%%%%%%%%%%%%%%%%%%%%%%%%%%%%%%%%%%%%%%
\begin{abstract}
Sewing garments using robots has consistently posed a research challenge due to the inherent complexities in fabric manipulation. In this paper, we introduce an intelligent robotic automation system designed to address this issue. By employing a patented technique that temporarily stiffens garments, we eliminate the traditional necessity for fabric modeling. Our methodological approach is rooted in a meticulously designed three-stage pipeline: first, an accurate pose estimation of the cut fabric pieces; second, a procedure to temporarily join fabric pieces; and third, a closed-loop visual servoing technique for the sewing process. Demonstrating versatility across various fabric types, our approach has been successfully validated in practical settings, notably with cotton material at the Bluewater Defense production line and denim material at Levi's research facility. The techniques described in this paper integrate robotic mechanisms with traditional sewing machines, devising a real-time sewing algorithm, and providing hands-on validation through a collaborative robot setup.
\end{abstract}

%%%%%%%%%%%%%%%%%%%%%%%%%%%%%%%%%%%%%%%%%%%%%%%%%%%%%%%%%%%%%%%%%%%%%%%%%%%%%%%%
\section{Introduction}
Robots have found extensive adoption in sectors like electronics and automotive manufacturing, yet their presence remains limited in apparel manufacturing. This can be attributed to the challenges the robotic system faces when manipulating soft and flexible textiles. Thus, apparel manufacturing largely relies on human workers, either to sew manually or to feed fabric pieces to semi-automated machines. Many of these manual tasks are repetitive, leading to worker burnout, frequent absences, and labor scarcities.  Moreover, the semi-automated machines \cite{Juki} used in the industry are predominantly designed for specific, isolated operations resulting in a form of rigid automation in the industry that lacks the flexibility to swiftly adjust to shifts in the production line. Thus, it becomes evident that there's a pressing need for flexible automation mechanisms in apparel manufacturing — a role that robots are well-suited to fulfill provided they can handle the flexible and limp nature of textiles.

% The recent pandemic further highlighted the fragility of such a labor-dependent system, pointing to potential interruptions and reduced productivity during labor crises.
% Approximately about 20 billion labor-hours a year are spent sewing clothing- equivalent to 28,900 average lifetimes and costing 54B dollars\cite{Placeholder}. The majority of these tasks are characterized by their repetitive and monotonous nature, which often leads to frequent worker absenteeism and labor shortages. Furthermore, the recent pandemic has showed the vulnerabilities and  risks associated with an over-reliance on human labor in operations, potentially resulting in disruptions and decreased manufacturing output during labor shortages.
\begin{figure}
\centering
\vspace{0.5cm}
\def\svgwidth{\columnwidth}
{\scriptsize 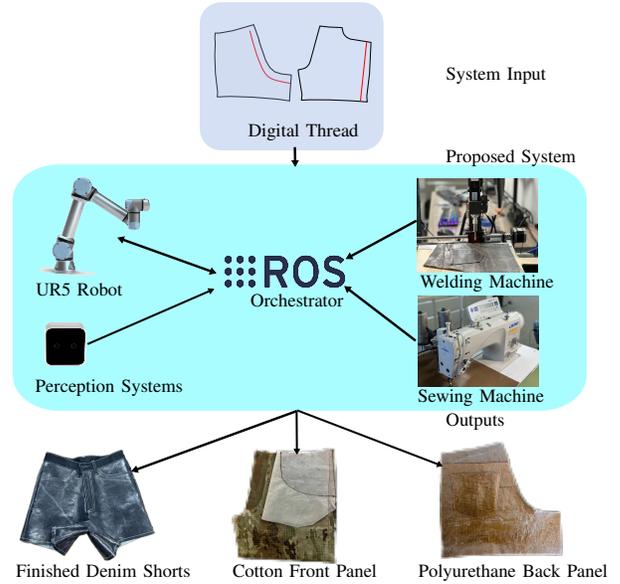}        \caption{\textbf{Sewing with Robots:} This figure provides a graphical representation of the robotic sewing system. The input for this system is a Drawing Interchange Format (DXF) supplied by the operator. The system's orchestration is managed through the Robot Operating System (ROS). Outputs of the system include various fabrics, including a cotton front panel, a polyurethane back panel, and the final product of finished denim shorts.}
    \label{fig:Trajectory}
\end{figure}
  In the past, numerous efforts have been made to automate the industry through the use of robots. In \cite{Schlegl2018RobotsIT}\cite{Nguyen2022}\cite{Jindal2021} the authors emphasize the various strategies that researchers and automation engineers have been diligently pursuing for years. However, the common thread among all these approaches is their heavy reliance on the intricate modeling of fabric properties, coupled with the advanced gripper design, complex vision and force sensing capabilities. To date, no fabric control strategy has successfully moved from the prototype stage to a tangible real-world application. Recognizing these practical challenges, we propose an approach designed for real-world application, ensuring seamless integration with existing manufacturing processes. Our approach to this fabric handling involves temporarily stiffening the fabric with a water-soluble "posing agent" that temporarily eliminates the limpness of the garment by stiffening it. The agent can be easily rinsed off and reused sustainably. This technology was developed and patented by Sewbo \cite{Sewbo}. This stiffened fabric allows us to utilize existing technologies developed for sheet handling, thereby eliminating the need to model fabric properties. Furthermore, once laminated, these garment pieces can be handled using a vacuum gripper removing the need for developing dexterous manipulation capabilities.  However, this approach has its challenges such as 
   \begin{enumerate}
  \item Environmental humidity can induce wrinkling or cause the garment piece to curl.
  \item Method to temporarily attach the pieces together before the sewing operation.
  \item Slippage of garment pieces from the vacuum gripper during the sewing operation.
\end{enumerate}

 To address the above-mentioned challenges, we have developed a novel three-stage pipeline to ensure the reliable sewing of pieces.  Initially, a robust pose estimation step identifies the position and orientation of garment pieces, taking into account potential fabric distortions like curling and wrinkling. The second step is to align the pieces together followed by a welding operation, similar to the ones used in the plastic industry. The final step is to sew the fused pieces reliably by performing closed-loop visual servoing of the robot and sewing machine. In implementing the outlined pipeline, we have achieved consistent success in sewing pieces across varied materials, shapes, and sizes. Moreover, the practical applicability of this technology has been showcased on the production floor of the Bluewater Defense line \cite{Garments}. Presently, our efforts are geared towards adapting this methodology to denim manufacturing, in a research venture with partners including Levi Strauss \& Co. \cite{BotCouture} and Saitex, a vertically integrated blue jeans manufacturer.
  We make 3 specific contributions in this work
  \begin{enumerate}
  \item We develop a robot sewing automation system which can be integrated with the conventional sewing machines in the apparel manufacturing industry with minimal disruptions.
  \item We develop a closed-loop sewing algorithm to monitor and perform robot sewing in real-time.
  \item We integrate our system with a collaborative robot and demonstrated it in a factory environment.
\end{enumerate}

\section{Related Work}
\subsection{Automated Sewing Systems}
In Koustoumpardis et al. \cite{KOUSTOUMPARDIS201434}, a higher-level controller trained on neural network estimates fabric properties and required tension for straight seams and sends it to the robot, but this approach needs a training data set to be created for the neural network and doesn't generalize well for new seam shape which is not part of the training data set. Lee et al. \cite{pr9020289} showcases an end-to-end sewing system with digital instructions, but requires custom-designed grippers and jigs. Schrimpf et al. \cite{6224880} employ two robots for sewing recliner covers but don't explore their efficacy on performing curved seams. In Schrimpf et al. \cite{Schrimpft2015}, authors introduced a multi-robot solution for 3D fabric sewing where they leveraged point-cloud based operations for pose estimation and tracking during sewing, yet its applicability to prevalent 2D sewing operations remain unaddressed. Winck et al. \cite{Winck2009} demonstrates open-loop sewing for arbitrary seams, yet highlights the necessity for closed-loop sewing in their discussion to address fabric slippage during sewing operation. In Kosaka et al. \cite{KOSAKA2023103005}, a closed-loop visual sewing system tracks the distance between the needle and edge for control, similar to our approach. However, they employ custom rollers for feeding, while we utilize suction-based end-effectors on the robot that makes our solution more adaptable to new operations and environments. Their system's constrained kinematic capability demands complex optimization for control decisions, whereas our 6-degrees of freedom arm simplifies the control problem and allows additional tasks like handling and transporting pieces within the operation.

Besides academia, industries offer solutions for automated sewing. Yaskawa's KMF Sewing robot \cite{MotomanIndustrialSewing} manufactures seat covers using specialized jigs, guiding the robot along a set path. Softwear Automation's Sewbots \cite{SoftWearAutomation} have custom machinery for T-shirts, executing all operations within one device; however, their cost and inflexibility limit their versatility. Robotextile \cite{RoboTextile} offers a unique end-effector to singulate garments from a pile and feed them to semi-automated machines. Pfaff Industrial's KL500 \cite{PfaffIndustrialRobotics} utilizes a 6-DOF arm with a custom sewing end-effector, catering to 3D sewing operations like seat covers.

\section{Methodology}
The robotic automation system we have developed encompasses several integral components: the Robot, the Sewing Machine, the Welding Tool, and the Camera System, all of which are illustrated in the corresponding Figure. \ref{fig:overview}.
\subsection{Digital Thread} 
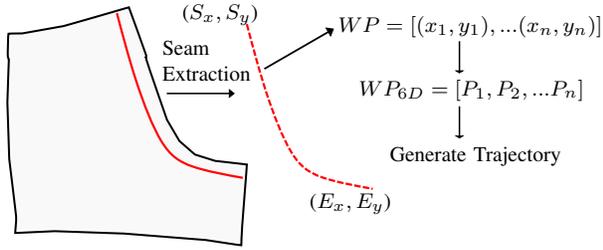
\begin{figure}
\centering
\vspace{.3cm}
\def\svgwidth{0.85\columnwidth}
{\footnotesize \input{DThreadTEstv3.pdf_tex}}    \caption{\textbf{Digital Thread:} Digital drawing of back panel of denim shorts that includes a seam designated in red by the designer. The figure also illustrates the extraction of the trajectory from this digital drawing.}
    \label{fig:DigitalThread}
\end{figure}

In apparel manufacturing, laser cutters employ Drawing Interchange Format (DXF) files for precise garment cutting. To enable seamless integration with this existing infrastructure, the proposed robotic system accepts the same digital drawings as input. Within these DXF files, the designer specifies the sewing seam to be traced using a designated color, for instance, red. Besides, data like seam allowance and stitch length will be incorporated in this file. This data is sent to the downstream pipeline tasks: Perception and Robot Trajectory Generation. The Perception module extracts the contour of the shape from these drawings and uses it to estimate the pose of the garments during the runtime. The robot trajectory generation module extracts  geometric entities such as lines and splines and generates trajectories based on them, adhering to velocity and acceleration constraints. An example digital drawing of back panel of denim shorts is shown in the Figure ~\ref{fig:DigitalThread}.

\subsection{Perception System}
The primary objective of the perception system is to estimate the pose $(x,y,\theta)$ of the workpiece on the workcell. To achieve this, a Realsense 435 Camera is positioned at 1.10 meters above the table, where it performs 2D pose estimation on the garment. The algorithm driving this process employs the same digital drawing as its input. The system also identifies the  "optimal grasp pixel" of the garment based on the garment dimension and the gripper constraints.

First, the algorithm employs a background subtraction method to distinguish the garment from its background. Subsequent steps involve edge and contour detection. The system then utilizes a template matching technique, using the template generated from the digital drawing as detailed in Section A, to identify the corresponding contour. From the matched template, the grasp pixel coordinates $(x,y)$ are provided in the pixel space which are then converted into robot frame using the 2D calibration technique. In addition to that, the yaw orientation $(\theta)$ of the piece with respect to the provided digital drawing is also calculated. Equipped with this data, the robot is then able to grasp the garment and place it on the welding table, where the subsequent welding process takes place. 
% After detecting edges and contours, we use the digital drawing from the previous step for template matching. The difficulty arises when the contours are distorted or discontinuous due to curling or warping of garments as shown in Fig-----. To tackle this, if template matching finds an area or contour with at least 80p match, it searches locally for corners to align with the nearest corresponding corner, ultimately returning the pose with the highest confidence score. This pose is then transformed in relation to the grasp pixel coordinate from the digital drawing, ultimately providing the precise pixel coordinates $(x,y)$ and the orientation $(\theta)$ required for the robot to effectively grasp the garment piece.
% The camera is calibrated with the Robot using 2D Calibration.This step transforms the pixel values and orientation data obtained from the camera into the robot's base frame coordinates. Equipped with this transformed data, the robot is then able to grasp the garment and place it on the welding table, where the subsequent welding process takes place. 

\subsection{Welding System}
\begin{figure*}[h]
    \vspace{0.3cm}
    \centering
    \def\svgwidth{0.82\textwidth}
    {\footnotesize 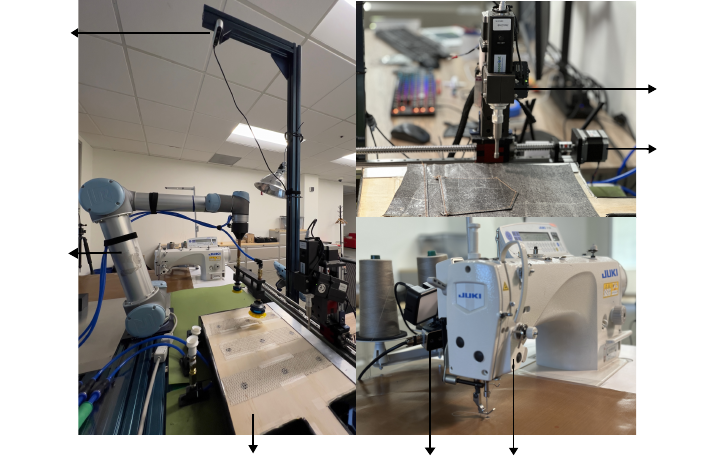}   
    \caption{\textbf{Overview of System Workcell}: (a) The first column provides a comprehensive view of the workcell, featuring the UR5 robot, welding table, sewing machine, and the overhead-mounted Realsense D435 camera for pose estimation. (b) The second column presents two images: the upper image showcases the ultrasonic welder mounted on the servo motors, whereas the lower image features the sewing machine utilized in our operation.}
    \label{fig:overview}
\end{figure*}

To facilitate the robot's ability to grasp and sew two pieces of fabric together, we employ ultrasonic welding as a preparatory step. This technique temporarily bonds the thermoplastic layers of the two pieces. An ultrasonic welder is employed for this purpose and is mounted on two stepper motors as shown in the Figure \ref{fig:overview}b. This allows precise control over the positioning of the welder, the garment piece, and enables welding operation at multiple spots. The operation of the ultrasonic welder is controlled by a dedicated low-level controller, which in turn is controlled by the main Robot Operating System (ROS) Controller. The number of welds required is determined by the length of the fabric pieces, while the duration of each individual weld is adjusted based on the thickness of the pieces.

\subsection{Sewing Machine Control}%
The sewing machine in our system is a JUKI CP-180 controller that is interfaced to a dedicated lower-level sewing machine microcontroller. This microcontroller serves as the interface between the machine's hardware and the overarching robotic system. It enables control over the presser foot, the trimming of the thread, and the overall enabling or disabling of the sewing machine. This microcontroller is integrated with Robot Operating System(ROS) to enable interoperability.

\subsection{Robot Trajectory Generation}
Robot control is orchestrated through the MoveIt! framework in ROS for pick-and-place operations, which include tasks like pose estimation and transferring the garment between machines such as welder and sewing machine, Cartesian trajectories are generated on a point-to-point basis within the defined planning scene. The robot follows these trajectories accordingly. 

However, this approach is not suitable for sewing trajectories. For sewing, the robot and the sewing
machine must operate at a constant speed to ensure consistent spacing between stitches. To achieve this, we generate the waypoints from the digital drawing as shown in Figure \ref{fig:DigitalThread}, followed by collision checking to ensure the waypoints are collision-free.  Given that MoveIt! does not natively support constant-velocity trajectories, we post-process the waypoints to generate  near-constant velocity trajectory adhering to the acceleration, velocity, and joint limit constraints. 

% Initially, a Forward Kinematic Solution is calculated between two waypoints in the given trajectory. Subsequently, the Euclidean distance between these points is computed. Time stamps are then determined based on a constant end-effector velocity. Using these new time stamps, joint velocities and accelerations are calculated. Constraints are assessed, and the trajectory is adjusted as needed to maintain a constant velocity throughout the operation.
%\begin{figure*}[!t]
%   \centering
%    \includegraphics[width=0.8\textwidth]{Images/SystemDiagram.png}
%    \caption{\textbf{System Architecture Overview}: The illustration delineates the comprehensive architecture, featuring both Open Loop and closed-loop mechanisms.Additionally, the diagram includes lower-level controllers for the sewing and welding machines and details the communication protocols among various controllers, perception nodes, and the robot. } 
%    \label{fig:overview}
%\end{figure*}

\subsection{Open-Loop Robot Sewing}%
For a uniform seam along the garment's edge, it's imperative that the robot and sewing machine work synchronously. Achieving this synchronization requires setting the constraints for the robot's velocity, acceleration, and joint limits according to the desired final stitch specifics. Key factors influencing these constraints include: 1) the seam allowance, 2) the maximum speed of the sewing needle, 3) stitch length, 4) material thickness, and 5) thread tension.

Before initiating the robotic sewing operation, the fabric pieces must first be welded as outlined in Section C. Once welded, the robot retrieves the unified garment pieces from the welding table and positions them on the sewing machine. Utilizing digital control, the sewing machine's presser foot is elevated, allowing the robot to align the garment precisely beneath the needle, adhering to the predetermined seam allowance. The robot then generates its sewing trajectory as described in the previous section. Subsequently, the presser foot is lowered, and both the robot and sewing machine commence their synchronized operation, following the defined parameters.

While the open-loop robotic sewing system offers a range of benefits—including ease of setup and straightforward control mechanisms in an ideal setting, it falls short when faced with challenges like imprecise pose estimation, fabric slippage, sewing curved seams, or handling thicker fabric pieces. 

\section{Closed-Loop Sewing}

In contrast to open-loop systems, closed-loop systems incorporate real-time feedback mechanisms. This enables the system to continuously adapt its behavior based on the current state, thereby providing a higher degree of control and accuracy. In the context of robotic sewing, this adaptability is particularly valuable, as it allows for real-time adjustments of the work piece, to ensure the seam distance from the edges of the garment is consistent throughout the stitch. One of the most critical components of our closed-loop system is the integration of a camera-based visual feedback system. The camera serves as the "eyes" of the robotic sewing operation, adeptly tracking garment edges in real-time. Using this sensory information and the digital drawing of the seam, the control signals for the robot are calculated in each time step.

\subsection{Sewing Machine Perception System}
A Realsense D405 camera is fixed to the sewing machine, with an unobstructed view of the sewing needle as shown in Figure \ref{fig:overview}b. This specific camera model was selected due to its effectiveness at close proximity. To ensure accurate edge detection, the camera is paired with a diffused light source, which effectively eliminates shadows. A Region of Interest (ROI) is defined from the center of the sewing needle. Within this ROI, critical edges are identified by detecting the outermost edges of the fabric. Our system employs two methods for edge detection: 1) the Traditional Computer Vision Method and 2) the Deep Learning-Based Method. The traditional approach utilizes Canny Edge Detection \cite{Canny1986}, favored for its efficiency and minimal computational requirements while maintaining high accuracy in edge detection. For the Canny Algorithm's threshold settings, which necessitate a minimum and maximum threshold for distinguishing between weak and strong edges, these thresholds are dynamically calculated in real-time. This calculation involves determining the median pixel intensity of the current image and adjusting the lower and upper bounds to be 50\% lower and higher, respectively, than the median intensity. These values are updated every second by analyzing all frames within that period, ensuring that the edge detection remains adaptive to changes in environmental lighting. Alternatively, we also tested the Holistically-Nested Edge Detection (HED) model \cite{HED}, where we trained the model on various fabric types and panel designs which offered superior accuracy and noise resistance but at the cost of higher computational demand and lower frame rates. So, for the purposes of this study, we conducted all experiments using the traditional Canny edge detection method due to its balance of performance and efficiency. The detection results are illustrated in Figure \ref{fig:closedloop}, where critical edges are highlighted in green and the edge nearest to the sewing needle is indicated by a red arrow, determined using a signed distance function.

A 6D calibration is conducted between the camera on the sewing machine and the robot to obtain the extrinsic matrix, which describes the spatial relationship between the camera and the robot's base frame. The edge closest to the sewing needle, as well as the pixel center of the needle itself, are transformed into the robot's base frame coordinates. The distance between these points is then calculated in meters (m) in the robot base frame.

\subsection{Closed-Loop Robot Sewing}
To perform corrections in real-time, the closed-loop control system should be able to perform real-time collision checking and generate collision-free velocities for the robot to move during every control cycle. Hence, we selected the MoveIt! Servo \cite{MoveitServo} package as our real-time controller. This package allows us to perform real-time corrections by sending velocity commands to the robot controller. Based on the velocity and the control loop cycle, PID settings of the controller should be fine-tuned to make precise, responsive, real-time corrections. 

\begin{figure}
\centering
\vspace{0.3cm}
\def\svgwidth{\columnwidth}
{\scriptsize 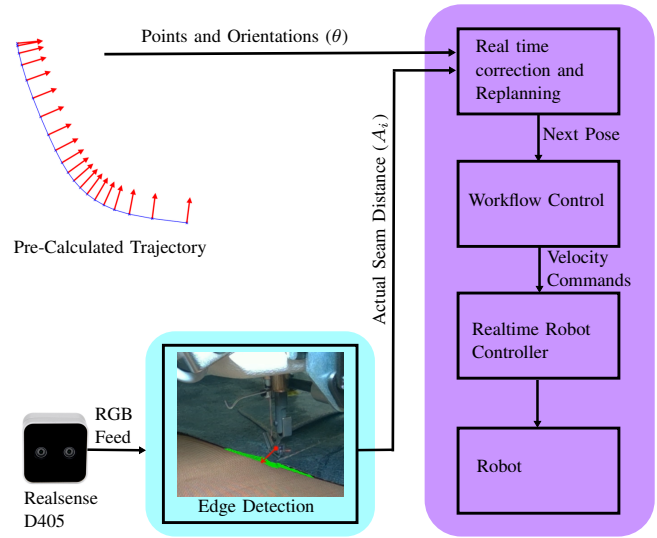}        \caption{\textbf{Block Diagram of Closed-Loop Control:}  The figure presents the schematic of the Integrated Closed-Loop Control framework.}
    \label{fig:closedloop}
\end{figure}

With the open-loop method, a trajectory is pre-planned before the start of the execution based on the digital drawing and corrections cannot be made to the trajectory during runtime. Therefore, here we propose an integrated approach that combines the real-time servo control with the  trajectory information from the Digital Thread along with garment edge data from the perception system as depicted in Figure \ref{fig:closedloop}. This allows for dynamic real-time adjustments, providing a control mechanism that's both adaptive and reactive.

The generated waypoints from the digital drawing are pre-processed in this step to keep them equidistant. Furthermore, the normals between each successive pair of waypoints are calculated, using the sewing machine as a reference point. These pre-calculated normals enable us to determine the orientation ($\theta$) of the edge with respect to the sewing needle, thereby providing valuable information for real-time adjustments.

\begin{algorithm}
\caption{Real-time Trajectory Correction and Replanning}
\label{alg:visualservoing}
\begin{algorithmic}[1]
\STATE \textbf{Input:} Trajectory, Seam tolerance (\( \text{tol} \)), Orientation \( \theta \)
\STATE \( x, y, \text{edgeDist}, \text{origin}, \text{tol}\) \hspace{1em} \(\triangleright\) Initialize variables
\STATE \( \text{origin} \leftarrow [0, 0] \) \hspace{5em} \(\triangleright\) Set origin
\STATE \( \text{corr} \leftarrow \text{False} \)\hspace{6em} \(\triangleright\) Initialize correction flag
\FOR{point in Trajectory} 
    \STATE \( x, y \leftarrow \text{point}[0], \text{point}[1] \) \(\triangleright\) Extract pixel coordinates
    \STATE Compute \( \text{edgeDist} \) \hspace{1em} \(\triangleright\) Compute distance to edge
    \IF{\( \text{edgeDist} \) not in \( \text{tol} \)}
        \STATE Correct \( x, y \) using \( \text{edgeDist} \) and \( \theta \) 
        \STATE \( \text{corr} \leftarrow \text{True} \)
        \STATE Update \( \text{origin} \) \hspace{3em} \(\triangleright\) Shift origin at deviation
    \ENDIF
    \IF{\( \text{corr} \)}
        \STATE Replan with \( \text{origin} \)  \hspace{1em} \(\triangleright\) Replan trajectory
        \STATE \( \text{corr} \leftarrow \text{False} \) \hspace{3em} \(\triangleright\) Reset correction flag
    \ENDIF
\ENDFOR
\STATE \textbf{Output:} Corrected Trajectory
\end{algorithmic}
\end{algorithm}

To begin the closed-loop sewing pipeline, the robot positions the garment underneath the sewing needle, which serves as the starting point for calculating the robot trajectory based on the digital drawing. After computing the robot trajectory, the state machine activates both the sewing machine and robot controllers. The robot controller sends the appropriate velocity to the robot to reach the waypoint within the control cycle, while the sewing machine controller issues the necessary pulses to execute the designated number of stitches. As the robot progresses along this planned path, its integrated visual system continuously monitors the edges of the garment to ensure they remain within the defined seam tolerance from the stitch. If the values exceed the set tolerance, the robot takes immediate corrective action as per Algorithm \ref{alg:visualservoing}. The corrective actions are executed as translation motions in the robot's $x$ and $y$ axes. These correction values are calculated by resolving the orientation($\theta$) values for each waypoint, extracted from the digital drawing, into their respective $x$- and $y$-components with $d$ being the value in meters (m) the robot has to move in that direction. The robot controller then executes this corrective action in the successive control cycle to bring the edges within the specified tolerance. 

\begin{equation}
X_{\text{Robot}} = x_{\text{traj}}  + X_{\text{Origin}} + x_{\text{correction}} \label{eq:1}
\end{equation}
\begin{equation}
x_{\text{correction}} = d \times \cos(\theta) \label{eq:2}
\end{equation}
\begin{equation}
Y_{\text{Robot}} = y_{\text{traj}}  + Y_{\text{Origin}} + y_{\text{correction}} \label{eq:3}
\end{equation}
\begin{equation}
y_{\text{correction}} = d \times \sin(\theta)\label{eq:4}
\end{equation}

Equations \eqref{eq:1} through \eqref{eq:4} are used by the robot controller to calculate the robot's position \(X_{\text{Robot}}\) and \(Y_{\text{Robot}}\) in each control cycle. In these equations, \(x_{\text{traj}}\) and \(y_{\text{traj}}\) represent the current points of the trajectory acquired from the predefined robot trajectory calculated from the digital drawing, while \(x_{\text{correction}}\) and \(y_{\text{correction}}\) denote any necessary corrections in meters. 

The term \(X_{\text{Origin}}\) initially corresponds to the robot's starting position; however, it is subject to updates if deviations occur, affecting both \(X_{\text{Origin}}\) and \(x_{\text{traj}}\). Such updates are crucial for re-planning the robot path, accounting for deviations caused by corrections. This realignment process is outlined in Algorithm \ref{alg:visualservoing}. The corrections are applied only if the deviations surpass a predefined tolerance value; otherwise, the correction terms remain zero, indicating that the edges are properly aligned.

\section {Experimental Verification}
In this section, we evaluate the performance of the developed system by conducting tests on 40 different fabric samples which are of the types: denim, cotton and polyurethane. The robotic system features a Universal Robot (UR5) and is controlled by a computer running Ubuntu 20.04, equipped with an Intel Xeon CPU E5-2640 v4 clocked at 2.40GHz. This computer operates the State Machine, serving as the primary workflow orchestrator. It is connected to a RealSense D405 camera that is attached to the sewing machine and also interfaces with the lower-level controllers for both the sewing and welding machines. For pose estimation, a separate RealSense D435 camera is mounted above the robot cell. This camera is connected to an Nvidia Jetson Xavier, which communicates with the computer via ROS Nodes.

\begin{figure}
\centering
\vspace{0.5cm}

\def\svgwidth{\columnwidth}
{\tiny 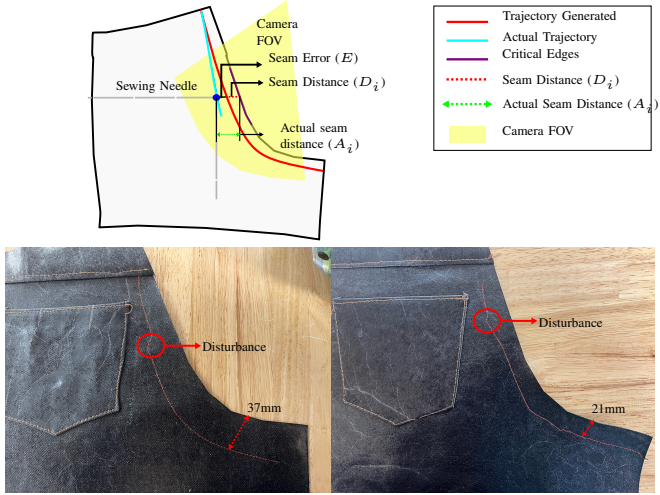}        \caption{\textbf{Closed-loop Results:} The top image serves as a reference illustrating the methodology for calculating seam error relative to seam distance and actual trajectory, with the Camera's Field of View (FOV) and the Sewing Needle remaining fixed. The bottom left image presents the outcomes of the open-loop system when subjected to disturbances. The bottom right image depicts the performance of the closed-loop system under identical conditions. \textit{(Fixed Seam Distance: 20mm)}}
    \label{fig:fabricexperiments}
\end{figure}

In our experiments, we focused on evaluating two key aspects of the robotic sewing system: 1) its efficiency with different seam shapes such as curves and lines, and 2) its ability to handle various fabric types such as denim, cotton and polyurethane. A critical measure of efficiency in our study is the seam error ($E$), which serves as a quantitative indicator of the system's accuracy and precision; this is visually detailed in Figure \ref{fig:fabricexperiments}. We conducted tests on both straight and curved seams by using the back and front panels of trousers, closely replicating actual manufacturing processes. Same seam shapes were chosen in both open-loop and closed-loop tests to compare their performances directly. The open-loop approach served as a baseline to benchmark against the closed-loop system's results. Additionally, we simulated fabric slippage by reducing the vacuum suction at the end-effector.

% We manually introduce disturbances to mimic potential slippage of the fabric from the end effector. We then repeat these tests using the closed-loop System. The results from both setups are subsequently compared for each type of seam to assess performance differences.

% Both the open-loop and closed-loop system perform similarly under ideal setting with zero disturbancesThe outcomes of the experiments are illustrated in Figure \ref{fig:experiments}, where both open-loop and closed-loop control methods were subjected to a minor disturbance. In the open-loop configuration, depicted on the left side of the figure, the trajectory deteriorates progressively following the disturbance. Conversely, in the closed-loop configuration shown on the right, the robot successfully compensates for the disturbance, returning the trajectory to within the predefined tolerance levels. 
The experimental results for denim pieces are presented in Figure \ref{fig:fabricexperiments}, with a comprehensive summary of all experimental runs provided in Table \ref{results}. Seam Error (\(E\)), as defined in Equation \ref{eq:5}, represents the average deviation between the actual (\(A_i\)) and intended (\(D_i\)) seam distances across the seam. This deviation is calculated for each 10mm segment of the seam's total length. Thus, \(n\) denotes the number of 10mm segments, determined by dividing the seam's total length by 10mm. The actual seam distance (\(A_i\)) for each segment is measured manually using a vernier caliper to ensure precise distance measurement between the stitch and the garment's edge.

\begin{equation}
E = \frac{1}{n} \sum_{i=1}^{n} \left| A_i - D_i \right| \label{eq:5}
\end{equation}

\begin{table}[h]
\centering
\caption{Summary of Loop Type Performance}
\label{results}
\begin{tabular}{|c|c|c|c|}
\hline
\textbf{Loop Type} & \textbf{Condition} & \textbf{Runs} & \textbf{Seam Error (mm)} \\
\hline
Open Loop & Without Disturbance & 10 & 1.37 \\
Open Loop & With Disturbance & 10 & 9.775 \\
Closed Loop & Without Disturbance & 10 & 1.585 \\
Closed Loop & With Disturbance & 10 & 3.11 \\
\hline
\end{tabular}
\end{table}

As per feedback from our industry collaborators in the apparel manufacturing sector, the accepted standard for seam error is less than 3mm, establishing it as a benchmark for quality control. Without disturbances, the open-loop system shows satisfactory performance but becomes significantly less reliable with disturbances. In contrast, the closed-loop system exhibits greater accuracy when faced with similar disturbances. Notably, in the absence of disturbances, the open-loop system performs slightly better, a phenomenon attributed to noise in the edge detection of the perception system causing minor inaccuracies.

% \begin{table}[h]
% \centering
% \caption{Summary of Loop Type Performance}
% \label{results}
% \begin{tabular}{|c|c|c|c|}
% \hline
% \textbf{Loop Type} & \textbf{Condition} & \textbf{Runs} & \textbf{Seam Error (mm)} \\
% \hline
% Open Loop & Without Disturbance & 10 & 1.37 \\
% Open Loop & With Disturbance & 10 & 9.775 \\
% Closed Loop & Without Disturbance & 10 & 1.585 \\
% Closed Loop & With Disturbance & 10 & 3.11 \\
% \hline
% \end{tabular}
% \end{table}

% \begin{table}[h]
% \centering
% \caption{Summary of Loop Type Performance}
% \label{mytable}
% \begin{tabularx}{\columnwidth}{|X|X|X|X|} % Adjust \columnwidth to fit your specific needs
% \hline
% \textbf{Loop} & \textbf{Condition} & \textbf{Fabric Sewn} & \textbf{Average Seam Error (mm)} \\ % Abbreviated headers
% \hline
% Open & Without Disturbance & 10 & 1.37 \\
% Open & With Disturbance & 10 & 9.775 \\
% Closed & Without Disturbance & 10 & 1.585 \\
% Closed & With Disturbance & 10 & 3.11 \\
% \hline
% \end{tabularx}
% \end{table}

\addtolength{\textheight}{-12cm}   % This command serves to balance the column lengths
                                  % on the last page of the document manually. It shortens
                                  % the textheight of the last page by a suitable amount.
                                  % This command does not take effect until the next page
                                  % so it should come on the page before the last. Make
                                  % sure that you do not shorten the textheight too much.

%%%%%%%%%%%%%%%%%%%%%%%%%%%%%%%%%%%%%%%%%%%%%%%%%%%%%%%%%%%%%%%%%%%%%%%%%%%%%%%%
\section {Conclusion}
This paper introduces a robotic automated sewing system capable of assembling and sewing any 2D seam on fabrics. We've established an end-to-end pipeline, which, given only the digital drawing of the fabric, can deduce and execute the assembly and sewing process. Furthermore, we've developed a visual feedback-driven closed-loop sewing control system, which can perform real-time detection and correction of deviations common in sewing operations. Experiments were conducted to assess the real-time performance of the implemented control algorithm. The outcomes indicate that it's feasible to produce seams that meet the quality benchmarks established by the apparel manufacturing sector using the proposed technique.

%%%%%%%%%%%%%%%%%%%%%%%%%%%%%%%%%%%%%%%%%%%%%%%%%%%%%%%%%%%%%%%%%%%%%%%%%%%%%%%%

%%%%%%%%%%%%%%%%%%%%%%%%%%%%%%%%%%%%%%%%%%%%%%%%%%%%%%%%%%%%%%%%%%%%%%%%%%%%%%%%

\section {ACKNOWLEDGMENT}
The Research was sponsored by the ARM (Advanced Robotics for Manufacturing) Institute through a grant from the Office of the Secretary of Defense and was accomplished under Agreement Number W911NF-17-3-0004. The views and conclusions contained in this document are those of the authors and should not be interpreted as representing the official policies, either expressed or implied, of the Office of the Secretary of Defense or the U.S. Government. The U.S. Government is authorized to reproduce and distribute reprints for Government purposes notwithstanding any copyright notation herein.

%%%%%%%%%%%%%%%%%%%%%%%%%%%%%%%%%%%%%%%%%%%%%%%%%%%%%%%%%%%%%%%%%%%%%%%%%%%%%%%%

\bibliographystyle{IEEEtran}
\bibliography{IEEEabrv,ref}

\end{document}

%% file: macros.tex
\usepackage{todonotes}

%% file: edited_first.pdf_tex
%% Creator: Inkscape 1.3 (1:1.3+202307231459+0e150ed6c4), www.inkscape.org
%% PDF/EPS/PS + LaTeX output extension by Johan Engelen, 2010
%% Accompanies image file 'edited_first.pdf' (pdf, eps, ps)
%%
%% To include the image in your LaTeX document, write
%%   \input{<filename>.pdf_tex}
%%  instead of
%%   \includegraphics{<filename>.pdf}
%% To scale the image, write
%%   \def\svgwidth{<desired width>}
%%   \input{<filename>.pdf_tex}
%%  instead of
%%   \includegraphics[width=<desired width>]{<filename>.pdf}
%%
%% Images with a different path to the parent latex file can
%% be accessed with the `import' package (which may need to be
%% installed) using
%%   \usepackage{import}
%% in the preamble, and then including the image with
%%   \import{<path to file>}{<filename>.pdf_tex}
%% Alternatively, one can specify
%%   \graphicspath{{<path to file>/}}
%% 
%% For more information, please see info/svg-inkscape on CTAN:
%%   http://tug.ctan.org/tex-archive/info/svg-inkscape
%%
\begingroup%
  \makeatletter%
  \providecommand\color[2][]{%
    \errmessage{(Inkscape) Color is used for the text in Inkscape, but the package 'color.sty' is not loaded}%
    \renewcommand\color[2][]{}%
  }%
  \providecommand\transparent[1]{%
    \errmessage{(Inkscape) Transparency is used (non-zero) for the text in Inkscape, but the package 'transparent.sty' is not loaded}%
    \renewcommand\transparent[1]{}%
  }%
  \providecommand\rotatebox[2]{#2}%
  \newcommand*\fsize{\dimexpr\f@size pt\relax}%
  \newcommand*\lineheight[1]{\fontsize{\fsize}{#1\fsize}\selectfont}%
  \ifx\svgwidth\undefined%
    \setlength{\unitlength}{241.99820709bp}%
    \ifx\svgscale\undefined%
      \relax%
    \else%
      \setlength{\unitlength}{\unitlength * \real{\svgscale}}%
    \fi%
  \else%
    \setlength{\unitlength}{\svgwidth}%
  \fi%
  \global\let\svgwidth\undefined%
  \global\let\svgscale\undefined%
  \makeatother%
  \begin{picture}(1,0.90766942)%
    \lineheight{1}%
    \setlength\tabcolsep{0pt}%
    \put(0,0){\includegraphics[width=\unitlength,page=1]{edited_first.pdf}}%
    \put(0.44138055,0.43316997){\color[rgb]{0,0,0}\makebox(0,0)[lt]{\lineheight{1.25}\smash{\begin{tabular}[t]{l}Orchestrator\end{tabular}}}}%
    \put(0.11053865,0.45019622){\color[rgb]{0,0,0}\makebox(0,0)[lt]{\lineheight{1.25}\smash{\begin{tabular}[t]{l}UR5 Robot\end{tabular}}}}%
    \put(0.10997302,0.30143473){\color[rgb]{0,0,0}\makebox(0,0)[lt]{\lineheight{1.25}\smash{\begin{tabular}[t]{l}Perception Systems\end{tabular}}}}%
    \put(0.69861382,0.28593863){\color[rgb]{0,0,0}\makebox(0,0)[lt]{\lineheight{1.25}\smash{\begin{tabular}[t]{l}Sewing Machine\end{tabular}}}}%
    \put(0.70230556,0.46259347){\color[rgb]{0,0,0}\makebox(0,0)[lt]{\lineheight{1.25}\smash{\begin{tabular}[t]{l}Welding Machine\end{tabular}}}}%
    \put(0,0){\includegraphics[width=\unitlength,page=2]{edited_first.pdf}}%
    \put(0.07996868,0.01630896){\color[rgb]{0,0,0}\makebox(0,0)[lt]{\lineheight{1.25}\smash{\begin{tabular}[t]{l}Finished Denim Shorts\end{tabular}}}}%
    \put(0.41348779,0.01630858){\color[rgb]{0,0,0}\makebox(0,0)[lt]{\lineheight{1.25}\smash{\begin{tabular}[t]{l}Cotton Front Panel\end{tabular}}}}%
    \put(0.69938592,0.01630867){\color[rgb]{0,0,0}\makebox(0,0)[lt]{\lineheight{1.25}\smash{\begin{tabular}[t]{l}Polyurethane Back Panel\end{tabular}}}}%
    \put(0.74200261,0.78181015){\color[rgb]{0,0,0}\makebox(0,0)[lt]{\lineheight{1.25}\smash{\begin{tabular}[t]{l}System Input\end{tabular}}}}%
    \put(0.74200265,0.65532173){\color[rgb]{0,0,0}\makebox(0,0)[lt]{\lineheight{1.25}\smash{\begin{tabular}[t]{l}Proposed System\end{tabular}}}}%
    \put(0.74200265,0.24874797){\color[rgb]{0,0,0}\makebox(0,0)[lt]{\lineheight{1.25}\smash{\begin{tabular}[t]{l}Outputs\end{tabular}}}}%
    \put(0,0){\includegraphics[width=\unitlength,page=3]{edited_first.pdf}}%
    \put(0.4374609,0.6954155){\color[rgb]{0,0,0}\makebox(0,0)[lt]{\lineheight{1.25}\smash{\begin{tabular}[t]{l}Digital Thread\end{tabular}}}}%
    \put(0,0){\includegraphics[width=\unitlength,page=4]{edited_first.pdf}}%
  \end{picture}%
\endgroup%

%% file: DThreadTestv3.pdf_tex
%% Creator: Inkscape 1.3 (1:1.3+202307231459+0e150ed6c4), www.inkscape.org
%% PDF/EPS/PS + LaTeX output extension by Johan Engelen, 2010
%% Accompanies image file 'DThreadTestv3.pdf' (pdf, eps, ps)
%%
%% To include the image in your LaTeX document, write
%%   \input{<filename>.pdf_tex}
%%  instead of
%%   \includegraphics{<filename>.pdf}
%% To scale the image, write
%%   \def\svgwidth{<desired width>}
%%   \input{<filename>.pdf_tex}
%%  instead of
%%   \includegraphics[width=<desired width>]{<filename>.pdf}
%%
%% Images with a different path to the parent latex file can
%% be accessed with the `import' package (which may need to be
%% installed) using
%%   \usepackage{import}
%% in the preamble, and then including the image with
%%   \import{<path to file>}{<filename>.pdf_tex}
%% Alternatively, one can specify
%%   \graphicspath{{<path to file>/}}
%% 
%% For more information, please see info/svg-inkscape on CTAN:
%%   http://tug.ctan.org/tex-archive/info/svg-inkscape
%%
\begingroup%
  \makeatletter%
  \providecommand\color[2][]{%
    \errmessage{(Inkscape) Color is used for the text in Inkscape, but the package 'color.sty' is not loaded}%
    \renewcommand\color[2][]{}%
  }%
  \providecommand\transparent[1]{%
    \errmessage{(Inkscape) Transparency is used (non-zero) for the text in Inkscape, but the package 'transparent.sty' is not loaded}%
    \renewcommand\transparent[1]{}%
  }%
  \providecommand\rotatebox[2]{#2}%
  \newcommand*\fsize{\dimexpr\f@size pt\relax}%
  \newcommand*\lineheight[1]{\fontsize{\fsize}{#1\fsize}\selectfont}%
  \ifx\svgwidth\undefined%
    \setlength{\unitlength}{242.56074524bp}%
    \ifx\svgscale\undefined%
      \relax%
    \else%
      \setlength{\unitlength}{\unitlength * \real{\svgscale}}%
    \fi%
  \else%
    \setlength{\unitlength}{\svgwidth}%
  \fi%
  \global\let\svgwidth\undefined%
  \global\let\svgscale\undefined%
  \makeatother%
  \begin{picture}(1,0.43482323)%
    \lineheight{1}%
    \setlength\tabcolsep{0pt}%
    \put(0,0){\includegraphics[width=\unitlength,page=1]{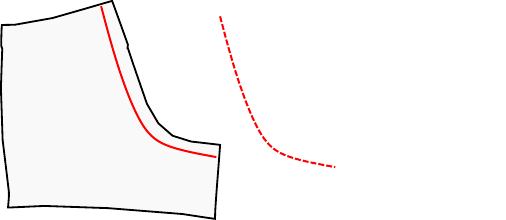}}%
    \put(0.31867088,0.41200916){\color[rgb]{0,0,0}\makebox(0,0)[lt]{\lineheight{1.25}\smash{\begin{tabular}[t]{l}($S_{x},S_{y}$)\end{tabular}}}}%
    \put(0.5495916,0.06921006){\color[rgb]{0,0,0}\makebox(0,0)[lt]{\lineheight{1.25}\smash{\begin{tabular}[t]{l}($E_{x},E_{y}$)\end{tabular}}}}%
    \put(0.28254654,0.34757148){\color[rgb]{0,0,0}\makebox(0,0)[lt]{\lineheight{1.25}\smash{\begin{tabular}[t]{l}Seam\\Extraction\end{tabular}}}}%
    \put(0,0){\includegraphics[width=\unitlength,page=2]{DThreadTestv3.pdf}}%
    \put(0.60100338,0.38842128){\color[rgb]{0,0,0}\makebox(0,0)[lt]{\lineheight{1.25}\smash{\begin{tabular}[t]{l}$WP = [(x_1,y_1),...(x_n,y_n)]$\end{tabular}}}}%
    \put(0.63848506,0.27367477){\color[rgb]{0,0,0}\makebox(0,0)[lt]{\lineheight{1.25}\smash{\begin{tabular}[t]{l}$WP_{6D} = [P_1,P_2,...P_n]$\end{tabular}}}}%
    \put(0.6947076,0.15445191){\color[rgb]{0,0,0}\makebox(0,0)[lt]{\lineheight{1.25}\smash{\begin{tabular}[t]{l}Generate Trajectory\end{tabular}}}}%
  \end{picture}%
\endgroup%

%% file: fullworkcell.pdf_tex
%% Creator: Inkscape 1.3 (1:1.3+202307231459+0e150ed6c4), www.inkscape.org
%% PDF/EPS/PS + LaTeX output extension by Johan Engelen, 2010
%% Accompanies image file 'fullworkcell.pdf' (pdf, eps, ps)
%%
%% To include the image in your LaTeX document, write
%%   \input{<filename>.pdf_tex}
%%  instead of
%%   \includegraphics{<filename>.pdf}
%% To scale the image, write
%%   \def\svgwidth{<desired width>}
%%   \input{<filename>.pdf_tex}
%%  instead of
%%   \includegraphics[width=<desired width>]{<filename>.pdf}
%%
%% Images with a different path to the parent latex file can
%% be accessed with the `import' package (which may need to be
%% installed) using
%%   \usepackage{import}
%% in the preamble, and then including the image with
%%   \import{<path to file>}{<filename>.pdf_tex}
%% Alternatively, one can specify
%%   \graphicspath{{<path to file>/}}
%% 
%% For more information, please see info/svg-inkscape on CTAN:
%%   http://tug.ctan.org/tex-archive/info/svg-inkscape
%%
\begingroup%
  \makeatletter%
  \providecommand\color[2][]{%
    \errmessage{(Inkscape) Color is used for the text in Inkscape, but the package 'color.sty' is not loaded}%
    \renewcommand\color[2][]{}%
  }%
  \providecommand\transparent[1]{%
    \errmessage{(Inkscape) Transparency is used (non-zero) for the text in Inkscape, but the package 'transparent.sty' is not loaded}%
    \renewcommand\transparent[1]{}%
  }%
  \providecommand\rotatebox[2]{#2}%
  \newcommand*\fsize{\dimexpr\f@size pt\relax}%
  \newcommand*\lineheight[1]{\fontsize{\fsize}{#1\fsize}\selectfont}%
  \ifx\svgwidth\undefined%
    \setlength{\unitlength}{344.0774231bp}%
    \ifx\svgscale\undefined%
      \relax%
    \else%
      \setlength{\unitlength}{\unitlength * \real{\svgscale}}%
    \fi%
  \else%
    \setlength{\unitlength}{\svgwidth}%
  \fi%
  \global\let\svgwidth\undefined%
  \global\let\svgscale\undefined%
  \makeatother%
  \begin{picture}(1,0.65148289)%
    \lineheight{1}%
    \setlength\tabcolsep{0pt}%
    \put(0,0){\includegraphics[width=\unitlength,page=1]{fullworkcell.pdf}}%
    \put(-0.00134017,0.29153056){\color[rgb]{0,0,0}\makebox(0,0)[lt]{\lineheight{1.25}\smash{\begin{tabular}[t]{l}UR5 Robot\end{tabular}}}}%
    \put(0.31711068,0.00151183){\color[rgb]{0,0,0}\makebox(0,0)[lt]{\lineheight{1.25}\smash{\begin{tabular}[t]{l}Pose Estimation Table\end{tabular}}}}%
    \put(0.53394734,0.00151183){\color[rgb]{0,0,0}\makebox(0,0)[lt]{\lineheight{1.25}\smash{\begin{tabular}[t]{l}Realsense D405\end{tabular}}}}%
    \put(0.67628429,0.00320737){\color[rgb]{0,0,0}\makebox(0,0)[lt]{\lineheight{1.25}\smash{\begin{tabular}[t]{l}Sewing Machine\end{tabular}}}}%
    \put(0.92355007,0.52935942){\color[rgb]{0,0,0}\makebox(0,0)[lt]{\lineheight{1.25}\smash{\begin{tabular}[t]{l}Ultrasonic\\Welder\end{tabular}}}}%
    \put(0.9225117,0.44763583){\color[rgb]{0,0,0}\makebox(0,0)[lt]{\lineheight{1.25}\smash{\begin{tabular}[t]{l}Servo\\Motor\end{tabular}}}}%
    \put(0.01609777,0.59984869){\color[rgb]{0,0,0}\makebox(0,0)[lt]{\lineheight{1.25}\smash{\begin{tabular}[t]{l}Realsense\\D435\end{tabular}}}}%
  \end{picture}%
\endgroup%

%% file: test.pdf_tex
%% Creator: Inkscape 1.3 (1:1.3+202307231459+0e150ed6c4), www.inkscape.org
%% PDF/EPS/PS + LaTeX output extension by Johan Engelen, 2010
%% Accompanies image file 'test.pdf' (pdf, eps, ps)
%%
%% To include the image in your LaTeX document, write
%%   \input{<filename>.pdf_tex}
%%  instead of
%%   \includegraphics{<filename>.pdf}
%% To scale the image, write
%%   \def\svgwidth{<desired width>}
%%   \input{<filename>.pdf_tex}
%%  instead of
%%   \includegraphics[width=<desired width>]{<filename>.pdf}
%%
%% Images with a different path to the parent latex file can
%% be accessed with the `import' package (which may need to be
%% installed) using
%%   \usepackage{import}
%% in the preamble, and then including the image with
%%   \import{<path to file>}{<filename>.pdf_tex}
%% Alternatively, one can specify
%%   \graphicspath{{<path to file>/}}
%% 
%% For more information, please see info/svg-inkscape on CTAN:
%%   http://tug.ctan.org/tex-archive/info/svg-inkscape
%%
\begingroup%
  \makeatletter%
  \providecommand\color[2][]{%
    \errmessage{(Inkscape) Color is used for the text in Inkscape, but the package 'color.sty' is not loaded}%
    \renewcommand\color[2][]{}%
  }%
  \providecommand\transparent[1]{%
    \errmessage{(Inkscape) Transparency is used (non-zero) for the text in Inkscape, but the package 'transparent.sty' is not loaded}%
    \renewcommand\transparent[1]{}%
  }%
  \providecommand\rotatebox[2]{#2}%
  \newcommand*\fsize{\dimexpr\f@size pt\relax}%
  \newcommand*\lineheight[1]{\fontsize{\fsize}{#1\fsize}\selectfont}%
  \ifx\svgwidth\undefined%
    \setlength{\unitlength}{244.09503174bp}%
    \ifx\svgscale\undefined%
      \relax%
    \else%
      \setlength{\unitlength}{\unitlength * \real{\svgscale}}%
    \fi%
  \else%
    \setlength{\unitlength}{\svgwidth}%
  \fi%
  \global\let\svgwidth\undefined%
  \global\let\svgscale\undefined%
  \makeatother%
  \begin{picture}(1,0.81994562)%
    \lineheight{1}%
    \setlength\tabcolsep{0pt}%
    \put(0,0){\includegraphics[width=\unitlength,page=1]{test.pdf}}%
    \put(0.83423955,0.421522){\color[rgb]{0,0,0}\makebox(0,0)[lt]{\lineheight{1.25}\smash{\begin{tabular}[t]{l}Velocity \\Commands\end{tabular}}}}%
    \put(0.82789714,0.6107297){\color[rgb]{0,0,0}\makebox(0,0)[lt]{\lineheight{1.25}\smash{\begin{tabular}[t]{l}Next Pose\end{tabular}}}}%
    \put(0,0){\includegraphics[width=\unitlength,page=2]{test.pdf}}%
    \put(0.20993422,0.76197048){\color[rgb]{0,0,0}\makebox(0,0)[lt]{\lineheight{1.25}\smash{\begin{tabular}[t]{l}Points and Orientations ($\theta$)\end{tabular}}}}%
    \put(0.58522253,0.32577261){\color[rgb]{0,0,0}\rotatebox{90}{\makebox(0,0)[lt]{\lineheight{1.25}\smash{\begin{tabular}[t]{l}Actual Seam Distance (\(A_i\))\end{tabular}}}}}%
    \put(0,0){\includegraphics[width=\unitlength,page=3]{test.pdf}}%
    \put(0.29711941,0.03779777){\color[rgb]{0,0,0}\makebox(0,0)[lt]{\lineheight{1.25}\smash{\begin{tabular}[t]{l}Edge Detection\end{tabular}}}}%
    \put(0,0){\includegraphics[width=\unitlength,page=4]{test.pdf}}%
    \put(0.72470921,0.74580546){\color[rgb]{0,0,0}\makebox(0,0)[lt]{\lineheight{1.25}\smash{\begin{tabular}[t]{l}Real time\\correction and \\Replanning\end{tabular}}}}%
    \put(0,0){\includegraphics[width=\unitlength,page=5]{test.pdf}}%
    \put(0.71352512,0.50863127){\color[rgb]{0,0,0}\makebox(0,0)[lt]{\lineheight{1.25}\smash{\begin{tabular}[t]{l}Workflow Control\end{tabular}}}}%
    \put(0.71913765,0.31051652){\color[rgb]{0,0,0}\makebox(0,0)[lt]{\lineheight{1.25}\smash{\begin{tabular}[t]{l}Realtime Robot \\Controller\end{tabular}}}}%
    \put(0,0){\includegraphics[width=\unitlength,page=6]{test.pdf}}%
    \put(0.72609663,0.09966027){\color[rgb]{0,0,0}\makebox(0,0)[lt]{\lineheight{1.25}\smash{\begin{tabular}[t]{l}Robot\end{tabular}}}}%
    \put(0,0){\includegraphics[width=\unitlength,page=7]{test.pdf}}%
    \put(0.02917426,0.04987877){\color[rgb]{0,0,0}\makebox(0,0)[lt]{\lineheight{1.25}\smash{\begin{tabular}[t]{l}Realsense \\D405\end{tabular}}}}%
    \put(0.01331813,0.43743186){\color[rgb]{0,0,0}\makebox(0,0)[lt]{\lineheight{1.25}\smash{\begin{tabular}[t]{l}Pre-Calculated Trajectory\end{tabular}}}}%
    \put(0,0){\includegraphics[width=\unitlength,page=8]{test.pdf}}%
    \put(0.1386955,0.17982541){\color[rgb]{0,0,0}\makebox(0,0)[lt]{\lineheight{1.25}\smash{\begin{tabular}[t]{l}RGB\\Feed\\\end{tabular}}}}%
  \end{picture}%
\endgroup%

%% file: expresults_edited3.pdf_tex
%% Creator: Inkscape 1.3 (1:1.3+202307231459+0e150ed6c4), www.inkscape.org
%% PDF/EPS/PS + LaTeX output extension by Johan Engelen, 2010
%% Accompanies image file 'expresults_edited3.pdf' (pdf, eps, ps)
%%
%% To include the image in your LaTeX document, write
%%   \input{<filename>.pdf_tex}
%%  instead of
%%   \includegraphics{<filename>.pdf}
%% To scale the image, write
%%   \def\svgwidth{<desired width>}
%%   \input{<filename>.pdf_tex}
%%  instead of
%%   \includegraphics[width=<desired width>]{<filename>.pdf}
%%
%% Images with a different path to the parent latex file can
%% be accessed with the `import' package (which may need to be
%% installed) using
%%   \usepackage{import}
%% in the preamble, and then including the image with
%%   \import{<path to file>}{<filename>.pdf_tex}
%% Alternatively, one can specify
%%   \graphicspath{{<path to file>/}}
%% 
%% For more information, please see info/svg-inkscape on CTAN:
%%   http://tug.ctan.org/tex-archive/info/svg-inkscape
%%
\begingroup%
  \makeatletter%
  \providecommand\color[2][]{%
    \errmessage{(Inkscape) Color is used for the text in Inkscape, but the package 'color.sty' is not loaded}%
    \renewcommand\color[2][]{}%
  }%
  \providecommand\transparent[1]{%
    \errmessage{(Inkscape) Transparency is used (non-zero) for the text in Inkscape, but the package 'transparent.sty' is not loaded}%
    \renewcommand\transparent[1]{}%
  }%
  \providecommand\rotatebox[2]{#2}%
  \newcommand*\fsize{\dimexpr\f@size pt\relax}%
  \newcommand*\lineheight[1]{\fontsize{\fsize}{#1\fsize}\selectfont}%
  \ifx\svgwidth\undefined%
    \setlength{\unitlength}{161.61210632bp}%
    \ifx\svgscale\undefined%
      \relax%
    \else%
      \setlength{\unitlength}{\unitlength * \real{\svgscale}}%
    \fi%
  \else%
    \setlength{\unitlength}{\svgwidth}%
  \fi%
  \global\let\svgwidth\undefined%
  \global\let\svgscale\undefined%
  \makeatother%
  \begin{picture}(1,0.7552996)%
    \lineheight{1}%
    \setlength\tabcolsep{0pt}%
    \put(0,0){\includegraphics[width=\unitlength,page=1]{expresults_edited3.pdf}}%
    \put(0.40546486,0.62579214){\color[rgb]{0,0,0}\makebox(0,0)[lt]{\lineheight{1.25}\smash{\begin{tabular}[t]{l}Seam Distance ($D_i$)\end{tabular}}}}%
    \put(0.40566652,0.66275797){\color[rgb]{0,0,0}\makebox(0,0)[lt]{\lineheight{1.25}\smash{\begin{tabular}[t]{l}Seam Error ($E$)\end{tabular}}}}%
    \put(0,0){\includegraphics[width=\unitlength,page=2]{expresults_edited3.pdf}}%
    \put(0.82140222,0.25543119){\color[rgb]{0,0,0}\makebox(0,0)[lt]{\lineheight{1.25}\smash{\begin{tabular}[t]{l}Disturbance\end{tabular}}}}%
    \put(0.30344455,0.21768708){\color[rgb]{0,0,0}\makebox(0,0)[lt]{\lineheight{1.25}\smash{\begin{tabular}[t]{l}Disturbance\end{tabular}}}}%
    \put(0.3731036,0.1246516){\color[rgb]{0,0,0}\makebox(0,0)[lt]{\lineheight{1.25}\smash{\begin{tabular}[t]{l}37mm\end{tabular}}}}%
    \put(0.90268319,0.12200807){\color[rgb]{0,0,0}\makebox(0,0)[lt]{\lineheight{1.25}\smash{\begin{tabular}[t]{l}21mm\end{tabular}}}}%
    \put(0,0){\includegraphics[width=\unitlength,page=3]{expresults_edited3.pdf}}%
    \put(0.38739249,0.71528726){\color[rgb]{0,0,0}\makebox(0,0)[lt]{\lineheight{1.25}\smash{\begin{tabular}[t]{l}Camera \\FOV\end{tabular}}}}%
    \put(0.76611912,0.7287684){\color[rgb]{0,0,0}\makebox(0,0)[lt]{\lineheight{1.25}\smash{\begin{tabular}[t]{l}Trajectory Generated \\\end{tabular}}}}%
    \put(0,0){\includegraphics[width=\unitlength,page=4]{expresults_edited3.pdf}}%
    \put(0.17045664,0.62006228){\color[rgb]{0,0,0}\makebox(0,0)[lt]{\lineheight{1.25}\smash{\begin{tabular}[t]{l}Sewing Needle\end{tabular}}}}%
    \put(0.4233937,0.55597187){\color[rgb]{0,0,0}\makebox(0,0)[lt]{\lineheight{1.25}\smash{\begin{tabular}[t]{l}Actual seam \\ distance ($A_i$)\end{tabular}}}}%
    \put(0.76603224,0.69593497){\color[rgb]{0,0,0}\makebox(0,0)[lt]{\lineheight{1.25}\smash{\begin{tabular}[t]{l}Actual Trajectory\\\end{tabular}}}}%
    \put(0.76564135,0.66896199){\color[rgb]{0,0,0}\makebox(0,0)[lt]{\lineheight{1.25}\smash{\begin{tabular}[t]{l}Critical Edges\\\end{tabular}}}}%
    \put(0.76556241,0.62892886){\color[rgb]{0,0,0}\makebox(0,0)[lt]{\lineheight{1.25}\smash{\begin{tabular}[t]{l}Seam Distance ($D_i$)\end{tabular}}}}%
    \put(0.76603224,0.5921563){\color[rgb]{0,0,0}\makebox(0,0)[lt]{\lineheight{1.25}\smash{\begin{tabular}[t]{l}Actual Seam Distance ($A_i$)\end{tabular}}}}%
    \put(0.76564135,0.55050574){\color[rgb]{0,0,0}\makebox(0,0)[lt]{\lineheight{1.25}\smash{\begin{tabular}[t]{l}Camera FOV\end{tabular}}}}%
  \end{picture}%
\endgroup%